\documentclass[sigconf]{acmart}

\usepackage{booktabs} % For formal tables
\usepackage{amsmath}
\usepackage{amssymb}
\usepackage{tablefootnote}
\DeclareMathOperator*{\argmax}{argmax}

\makeatletter
\def\@fnsymbol#1{\ensuremath{\ifcase#1\or \dagger\or *\else\@ctrerr\fi}}
\makeatother

\begin{document}
\title{Learning User Preferences and Understanding Calendar Contexts for Event Scheduling}

\author{Donghyeon Kim}
\authornote{Both authors contributed equally to this work.}
\orcid{0000-0002-8224-8354}
\affiliation{
  \institution{Korea University}
  \city{Seoul}
  \postcode{02841}
  \country{Republic of Korea}
}
\email{donghyeon@korea.ac.kr}

\author{Jinhyuk Lee}
\authornotemark[1]
\orcid{0000-0003-4972-239X}
\affiliation{
  \institution{Korea University}
  \city{Seoul} 
  \postcode{02841}
  \country{Republic of Korea}
}
\email{jinhyuk\_lee@korea.ac.kr}

\author{Donghee Choi}
\affiliation{
  \institution{Korea University}
  \city{Seoul}
  \postcode{02841}
  \country{Republic of Korea}
}
\email{choidonghee@korea.ac.kr}

\author{Jaehoon Choi}
\affiliation{
  \institution{Konolabs, Inc.}
  \city{Seoul} 
  \postcode{02841}
  \country{Republic of Korea}
}
\email{jchoi@kono.ai}

\author{Jaewoo Kang}
\authornote{Corresponding author.}
\affiliation{
  \institution{Korea University}
  \city{Seoul}
  \postcode{02841}
  \country{Republic of Korea}
}
\email{kangj@korea.ac.kr}

% \renewcommand{\shortauthors}{D. Kim and J. Lee et al.}

\begin{abstract}
With online calendar services gaining popularity worldwide, calendar data has become one of the richest context sources for understanding human behavior. However, event scheduling is still time-consuming even with the development of online calendars. Although machine learning based event scheduling models have automated scheduling processes to some extent, they often fail to understand subtle user preferences and complex calendar contexts with event titles written in natural language. In this paper, we propose Neural Event Scheduling Assistant (NESA) which learns user preferences and understands calendar contexts, directly from raw online calendars for fully automated and highly effective event scheduling. We leverage over 593K calendar events for NESA to learn scheduling personal events, and we further utilize NESA for multi-attendee event scheduling. NESA successfully incorporates deep neural networks such as Bidirectional Long Short-Term Memory, Convolutional Neural Network, and Highway Network for learning the preferences of each user and understanding calendar context based on natural languages. The experimental results show that NESA significantly outperforms previous baseline models in terms of various evaluation metrics on both personal and multi-attendee event scheduling tasks. Our qualitative analysis demonstrates the effectiveness of each layer in NESA and learned user preferences.
\end{abstract}



\copyrightyear{2018} 
\acmYear{2018} 
\setcopyright{none}
% \acmConference[CIKM '18]{The 27th ACM International Conference on Information and Knowledge Management}{October 22--26, 2018}{Torino, Italy}
% \acmBooktitle{The 27th ACM International Conference on Information and Knowledge Management (CIKM '18), October 22--26, 2018, Torino, Italy}
% \acmPrice{15.00}
% \acmDOI{10.1145/3269206.3271712}
% \acmISBN{978-1-4503-6014-2/18/10}


\maketitle

\section{Introduction}

\begin{figure}[ht]
\includegraphics[width=8.6cm,height=5.0cm]{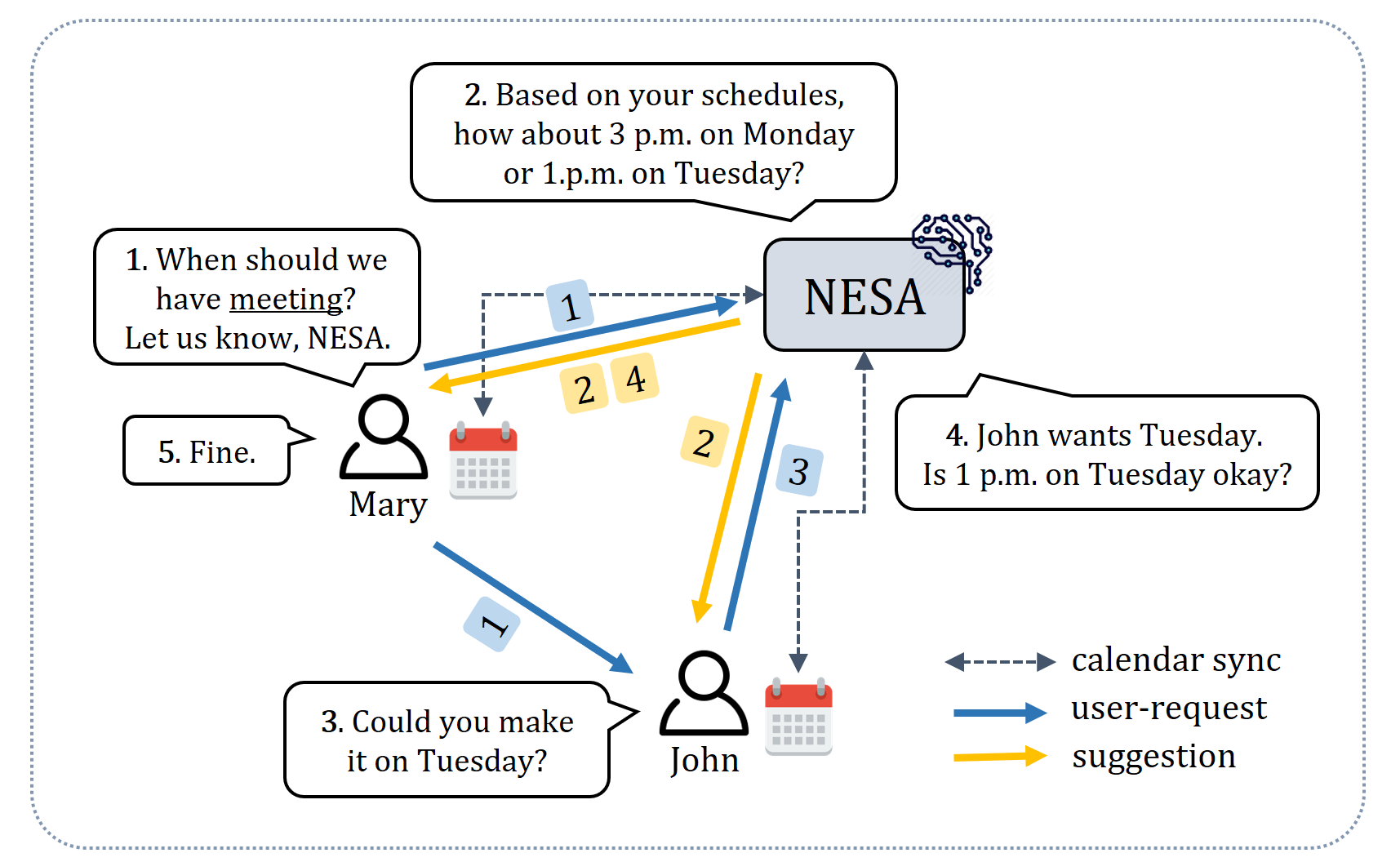}
\caption{Example of calendar event scheduling. Mary requests NESA to schedule a meeting with John. NESA considers each user's preference and calendar context, and the purpose of the event.}
\label{fig:task_intro}
\end{figure}

Calendar data has become an important context source of user information due to the popularity of online calendar services such as Google Calendar and Outlook Calendar. According to a research study conducted by Promotional Products Association International in 2011, about 40\% of people referred to calendars on their computers, and about 22\% of people used their mobile calendars every day \cite{saritha2011an}. As more people use online calendar services, more detailed user information is becoming available \cite{montoya2016thymeflow}.\\
\indent Event scheduling is one of the most common applications that uses calendar data \cite{berry2004personalized,blum1997empirical}. Similar to Gervasio et al. \cite{gervasio2005active} and Berry et al. \cite{berry2007balancing}, we define event scheduling as suggesting suitable time slots for calendar events given user preferences and calendar contexts. However, even with the development of communication technology, event scheduling is still time-consuming. According to Konolabs, Inc., the average number of emails sent between people to set the time for a meeting is 5.7.\footnote{Statistics obtained by Konolabs, Inc. (https://kono.ai) in 2017.} At the same time, the market for digital assistants is growing fast. Gartner, Inc. stated that by 2019, at least 25\% of households will use digital assistants on mobiles or other devices as primary interfaces of connected home services \cite{sun2016contextual}. Thus, it is important for digital assistants to effectively schedule users' events \cite{bjellerup2010falcon}.\\
\indent An example of scheduling an event using NESA is illustrated in Figure \ref{fig:task_intro}. When a user (\texttt{Mary}) requests NESA to arrange an appointment with the other user (\texttt{John}), NESA suggests candidate time slots considering the purpose of the event (e.g., \texttt{meeting}), preferences of each user (e.g., \texttt{Mary} usually has meetings in the afternoon, and \texttt{John} likes to have meetings early in the week), and each user's calendar context. As a result, NESA reduces the communication cost between the users by assisting with event scheduling.\\ % example: 1) reduce calendar checking time, 2) immediately response 24/7 
\indent Despite its importance, automated event scheduling \cite{mitchell1994experience,blum1997empirical,berry2004personalized} has had limited success due to several reasons. First, previous studies heavily relied on hand-crafted event features such as predefined event types, fixed office/lunch hours, and so on. In addition to the cost of defining the hand-crafted event features, they could not accurately understand calendar contexts based on natural language. For instance, if a user requests to schedule a \texttt{late lunch} with other users, traditional scheduling systems do not suggest late lunch hours unless the keyword \texttt{late} is registered in the systems. Furthermore, raw online calendars frequently contain abbreviations (e.g., \texttt{Mtg} stands for \texttt{meeting}) and misspellings. To deal with natural language, recent studies have combined human labor with scheduling systems \cite{cranshaw2017calendar}. Second, most previous studies have developed their own scheduling systems to learn user preferences, which makes it difficult to apply their methodologies to other scheduling systems. Despite the wide use of the Internet standard format iCalendar \cite{desruisseaux2009internet}, developing scheduling assistants based on iCalendar gained much less attention among researchers \cite{wainer2007scheduling}.\\
% * <susanniekim@gmail.com> 2018-05-22T17:58:11.747Z:
% 
% >  gained much less attention  
% uc and awkward  why did it gain less atten? these two ideas are unrelated 
% 
% ^ <dweller92@naver.com> 2018-05-23T06:25:33.536Z:
% 
% they are widely used in applications, but researches were not fully make use of the format.
%
% ^.
% * <susanniekim@gmail.com> 2018-05-22T17:50:45.936Z:
% 
% > contexts based on natural languages uc contexts with titles in nat lang?
% contexts are made up of events, and events contain titles, and titles are in nat lang. So we say that contexts are based on natural language. ,<-- uc
% 
% 
% ^.
% * <susanniekim@gmail.com> 2018-05-21T15:08:07.787Z:
% 
% > based on natural languages
% this part is uc contexts in nat lang? or calendar data in nat lang?
% 
% 
% 
% ^ <dweller92@naver.com> 2018-05-22T13:46:54.734Z:
% 
% contexts are made up of events, and events contain titles, and titles are in nat lang. So we say that contexts are based on natural language.
% 
% ^ <dweller92@naver.com> 2018-05-22T14:40:19.804Z.
% * <susanniekim@gmail.com> 2018-05-21T15:02:59.687Z:
% 
% > developed
% or designed?
% 
% ^ <dweller92@naver.com> 2018-05-22T13:47:21.497Z:
% 
% developed
%
% ^ <dweller92@naver.com> 2018-05-22T14:40:21.490Z.
\indent In this paper, we propose Neural Event Scheduling Assistant (NESA) which is a deep neural model that learns to schedule calendar events using raw user calendar data. NESA is a fully automated event scheduling assistant which learns user preferences and understands raw calendar contexts that include natural language. To understand various types of information in calendars, NESA leverages several deep neural networks such as Bidirectional Long Short-Term Memory (Bi-LSTM) \cite{hochreiter1997long,schuster1997bidirectional} and Convolutional Neural Network (CNN) \cite{krizhevsky2012imagenet}. The following four layers in NESA are jointly trained to schedule personal calendar events: 1) Title layer, 2) Intention layer, 3) Context layer, and 4) Output layer. After training, NESA is utilized for scheduling personal events (e.g., homework) and multi-attendee events (e.g., meetings). We compare NESA with previous preference learning models for event scheduling \cite{berry2011ptime,mitchell1994experience,gervasio2005active}, and find that NESA achieves the best performance in terms of various evaluation metrics.\\
\indent The contributions of our paper are four-fold.
\begin{itemize}
\item We introduce NESA, a fully automated event scheduling model, which learns user preferences and understands calendar contexts, directly from their raw calendar data.
\item NESA successfully incorporates deep neural networks for event scheduling tasks.
\item We train NESA on 593,207 real online calendar events in Internet standard format which is applicable to any calendar systems.
\item NESA achieves the best performance on both personal and multi-attendee event scheduling tasks compared with other preference learning models.
% * <susanniekim@gmail.com> 2018-05-22T18:07:42.063Z:
% 
% > preference learning models
% same as above
% 
% ^.
% * <susanniekim@gmail.com> 2018-05-18T18:05:30.228Z:
% 
% > preference learning
% event sched models?
% 
% ^ <krth32@gmail.com> 2018-05-20T07:02:59.728Z:
% 
% modified to "user preference learning"
%
% ^ <susanniekim@gmail.com> 2018-05-21T15:19:40.059Z:
% 
% same as above comment
%
% ^ <dweller92@naver.com> 2018-05-22T13:53:53.325Z:
% 
% deleted user , but added event scheduling.
%
% ^ <dweller92@naver.com> 2018-05-22T14:41:06.639Z.
\end{itemize}

\indent The rest of this paper is organized as follows. In Section 2, we introduce some related studies on event scheduling, and briefly discuss the recent rise of neural networks. In Section 3, we formulate personal and multi-attendee event scheduling tasks. In Section 4, we discuss our deep neural model NESA that consists of Bi-LSTM, CNN, and Highway Network. In Section 5, we introduce our dataset used for the event scheduling task and discuss our qualitative analysis along with the experimental results. We conclude the paper and discuss future work in Section 6. We make our source code and pretrained NESA\footnote{https://github.com/donghyeonk/nesa} available so that researchers and machine learning practitioners can easily apply NESA to their scheduling systems.

\section{RELATED WORK AND BACKGROUND}
\subsection{Preference Learning for Event Scheduling}
Since the development of online calendars, researchers have focused on learning user preferences for scheduling calendar events. Mitchell et al. proposed Calendar Apprentice (CAP) which is a decision tree based calendar manager that can learn user scheduling preferences from experience \cite{mitchell1994experience}. Blum et al. introduced the Winnow and weighted-majority algorithms that outperformed CAP \cite{blum1997empirical} on predicting various attributes of calendar events. Mynatt et al. also utilized the context of a user's calendar to infer the user's event attendance \cite{mynatt2001inferring}. Berry et al. proposed an assistant called Personalized Calendar Assistant (PCalM), which is based on Naive Bayesian, for ranking candidate schedules \cite{berry2004personalized}. Refanidis et al. have developed an intelligent calendar assistant which uses a hierarchical preference model \cite{refanidis2009scheduling}.\\ 
\indent However, most event scheduling models were based on specific calendar systems using hand-crafted event features such as predefined event types and system dependent features. Previous scheduling methodologies are rarely used for modern event scheduling systems due to the high cost of designing hand-crafted features. Also, it is difficult for existing models to understand user calendars that often include user written texts such as event titles. In this paper, we propose NESA which learns to schedule calendar events, directly using raw calendar data that contains natural language texts. As NESA is trained on the Internet standard format, it is generally applicable to other calendar systems.

\subsection{Multi-Attendee Event Scheduling}
Event scheduling has also been studied in the context of multi-attendee event scheduling. Researches on event scheduling focus on solving constraint satisfaction problems (CSPs), and such researches often assume that user preferences are already given. Garrido et al. used heuristics functions for finding the priority value of each available time interval \cite{garrido1996multi}. Wainer et al. proposed a model to find optimal time intervals based on user preferences and dealt with privacy issues of shared calendars \cite{wainer2007scheduling}. Zunino et al. developed Chronos, a multi-attendee meeting scheduling system that employs a Bayesian network to learn user preferences \cite{zunino2009chronos}. \\
\indent However, most multi-attendee event scheduling models still depend on their own scheduling systems. Furthermore, due to the small amount of existing calendar event data (e.g., 2K events of 2 users \cite{mitchell1994experience,blum1997empirical,zunino2009chronos}), some of the previous studies \cite{blum1997empirical, garrido1996multi} use complicated heuristic functions based on system dependent features to find proper time intervals, making their methodologies difficult to adopt. In contrast, NESA leverages 593K standard formatted events and learns event scheduling directly from raw calendar data. While the recent work of Cranshaw et al. relied on human labor for more effective event scheduling \cite{cranshaw2017calendar}, our event scheduling assistant is fully automated. We also demonstrate the effectiveness of NESA on multi-attendee event scheduling.

\subsection{Representation Learning using Deep Neural Networks}
Many classification tasks such as image classification \cite{krizhevsky2012imagenet}, sentiment analysis \cite{dos2014deep}, and named-entity recognition \cite{DBLP:journals/corr/LampleBSKD16} have benefited from the recent rise of neural networks. Deep neural networks learn how to represent raw inputs such as image pixels for any targeted task. Given a raw user calendar, NESA learns how to represent user preferences and calendar contexts for event scheduling. While the preliminary work of Mitchell et al. showed that decision tree based models with hand-crafted features are better than artificial neural network (ANN) based models with hand-crafted features \cite{mitchell1994experience}, our work is the first to show that deep neural networks are effective for event scheduling tasks with raw calendar data.\\
% * <krth32@gmail.com> 2018-05-20T08:39:19.327Z:
% 
% > produce comparable accuracy
% modified from "are superior"
% 
% ^ <susanniekim@gmail.com> 2018-05-21T16:35:24.615Z:
% 
% check again
%
% ^ <dweller92@naver.com> 2018-05-22T14:11:09.378Z:
% 
% meaning wrongly conveyed here. decision tree was better than ANN when both are combined with hand-crafted features.
%
% ^ <susanniekim@gmail.com> 2018-05-22T18:26:24.961Z:
% 
% specify  both -->  dec models with hand outperform ANN models with hand...?
% outperform? 
%
% ^ <dweller92@naver.com> 2018-05-23T06:29:34.394Z:
% 
% okay.
%
% ^.
% * <susanniekim@gmail.com> 2018-05-18T19:31:51.871Z:
% 
% > when combined 
% refers to decision or ANN?
% 
% ^ <krth32@gmail.com> 2018-05-20T08:35:44.669Z:
% 
% decision. modified to "when the decision tree based model is combined"
%
% ^ <susanniekim@gmail.com> 2018-05-21T16:39:06.814Z:
% 
% check
%
% ^ <dweller92@naver.com> 2018-05-22T14:11:33.621Z.
\indent Among various kinds of neural networks, Recurrent Neural Networks (RNNs) have achieved remarkable performance on natural-language processing (NLP) tasks such as language modeling \cite{mikolov2010recurrent}, machine translation \cite{DBLP:journals/corr/BahdanauCB14}, and so on. Inspired by a character-level language model \cite{kim2016character} and state-of-the-art question answering models \cite{seo2016bidirectional}, NESA handles various semantics coming from raw calendar events based on natural language. We use RNN and CNN to effectively represent user written event titles, and use Highway Network \cite{srivastava2015highway} to find nonlinear relationships among various calendar attributes.
% * <susanniekim@gmail.com> 2018-05-18T19:42:49.292Z:
% 
% >  to handle variations coming from raw calendar events based on natural languages.
% uc QA models that handle? 
% variations coming from raw calendar events based on natural languages is uc 
% 
% ^ <krth32@gmail.com> 2018-05-20T08:59:27.521Z:
% 
% *** let me check
%
% ^ <krth32@gmail.com> 2018-05-20T15:39:41.821Z:
% 
% *** modified to "that handle variations in semantics coming from raw calendar events based on natural languages."
% 
% ^ <susanniekim@gmail.com> 2018-05-21T16:51:39.224Z:
% 
% various semantics of raw cal events in nat lang?
%
% ^ <dweller92@naver.com> 2018-05-22T14:12:14.739Z:
% 
% yes. various semantics, but changed that=> to, because model is inspired by those char/QA models 'to' handle various semantics. QA models have nothing to do with calendar data.
% 
% ^ <susanniekim@gmail.com> 2018-05-22T18:33:01.539Z:
% 
% "to handle various semantics coming from raw calendar events based on natural languages is uc 
% Our model, inspired by ....., handles various semantics of raw cal events in nat lang....
%  
%
% ^ <dweller92@naver.com> 2018-05-23T06:30:36.066Z:
% 
% okay.
%
% ^.

\section{Problem Formulation}
\subsection{Attributes of Calendar Data}
A user's calendar data consists of sequences of events which are sorted by their registered time. Each calendar event has at least five attributes: (1) \textit{title} (what to do), (2) \textit{start time}, (3) \textit{duration}, (4) \textit{registered time}, and (5) \textit{user identifier} of an event. Although many other attributes (e.g., \textit{location, description}) exist, we focus on the most common attributes of events. Note that the title of each event in iCalendar format does not have a label that indicates the event type, whereas previous scheduling systems rely on a predefined set of event types.\\
\indent To simplify the problem, we group all the events of each user by the week in which their events start. For example, user A's events that start within the 15th week of 2018 will be grouped in A\_2018\_15. In each group, events are sorted by their registered time. For each user, all the $K$ events in a specific week can be expressed as follows: $E$ = {$e_{1}$, \dots, $e_{K}$}, and $e_{i}$ = ($x_{i}$, $t_{i}$, $d_{i}$, $u_{i}$) for $i=1$ to $K$ where $x_{i}$ indicates the start time, $t_{i}$ is the title, $d_{i}$ is the duration, and $u_{i}$ is the user identifier of $e_{i}$. We assume that $u_i$ represents the \textit{preference} of a user, $t_i$ and $d_i$ represent the \textit{purpose} of $i$-th event, and ${e_1, \dots, e_{i-1}}$ represent the \textit{context} of $i$-th event. Note that the context can be extended to multiple weeks.
% * <susanniekim@gmail.com> 2018-05-18T19:56:26.505Z:
% 
% >  ${e_1, \dots, e_{i-1}}$ represent
% e1 and ei ? 
% 
% ^ <krth32@gmail.com> 2018-05-20T09:10:23.393Z:
% 
% e_1 and e_{i-1} are right because the range is previous indexes of i 
%
% ^ <susanniekim@gmail.com> 2018-05-21T16:54:08.046Z:
% 
% e and e  represent? or e ....e represents? 
%
% ^ <dweller92@naver.com> 2018-05-22T14:16:23.555Z:
% 
% these are plural but would be verbose and inconsistent to say e,..and e represent. so keep it as it is.
% 
% ^ <dweller92@naver.com> 2018-05-22T14:43:20.076Z.
\subsection{Personal Event Scheduling}
\indent Event scheduling involves considering users' preferences and calendar contexts to provide suitable time slots to users. We define personal event scheduling as scheduling events that have a single attendee (e.g., work, personal matters, and so on). We later describe how to extend personal event scheduling to multi-attendee event scheduling.\\
\indent Personal event scheduling should consider the pre-registered events of the week (context) in which an event will be registered and the preferences of a user. Thus, an event scheduling model predicts the start time $y_i$ of the $i$-th event $e_{i}$ given the pre-registered events ($e_{1}$, \dots, $e_{i-1}$) which constitute the context of the week, and given the title $t_{i}$, duration $d_{i}$, and user $u_{i}$ attributes of the $i$-th event. Note that each pre-registered event also contains title, duration, user, and start time ($x_{i}$) attributes, making it difficult for any models to leverage all the given contexts.\\
% * <susanniekim@gmail.com> 2018-05-22T18:41:08.760Z:
% 
% >  a model
% what type of model?
% 
% ^ <dweller92@naver.com> 2018-05-23T06:31:43.965Z:
% 
% just any model. event scheduling model.
%
% ^.
% * <susanniekim@gmail.com> 2018-05-18T20:20:42.552Z:
% 
% > the model 
% sp 
% 
% ^ <krth32@gmail.com> 2018-05-20T09:11:57.644Z:
% 
% how about "NESA personal model"?
%
% ^ <susanniekim@gmail.com> 2018-05-21T16:55:21.477Z:
% 
% NESA?   it's best not to give it different names. 
% 
% ^ <dweller92@naver.com> 2018-05-22T14:17:13.543Z:
% 
% this should indicate any general models including baseline models. so 'a model'.
% 
% ^ <dweller92@naver.com> 2018-05-22T14:43:32.611Z.
\indent Given the probability of target time slot $y_i$ of event $e_{i}$, the optimal model parameters $\Theta^*$ are as follows:
\begin{equation}
\label{eq:optimize_params}
\Theta^* = \argmax_\Theta p(y_i|e_{1}, \dots, e_{i-1}, t_{i}, d_{i}, u_{i}; \Theta)
\end{equation}

\noindent where $\Theta$ denotes the trainable parameters of a model. Note that there exist K event scheduling problems in a week including weeks with no pre-registered events. We treat each event scheduling problem as an independent problem to measure the ability of each model to understand calendar contexts and user preferences.

\subsection{Multi-Attendee Event Scheduling}
Multi-attendee event scheduling further considers the preferences and calendar contexts of multiple users attending an event. Given $U$ users attending a specific event $e_{\mu}$ with the optimal model parameter $\Theta^*$, the most suitable time slot $y_{\mu}^*$ among candidate time slots $\hat{y}_{\mu}$ is computed as follows:
\begin{equation}
y_{\mu}^* = \argmax_{\hat{y}_{\mu}} \sum_{j=1}^U p(\hat{y}_{\mu}|E_{1:\mu-1}^j,t_{\mu},d_{\mu},u_j;\Theta^*)
\label{eq:multi}
\end{equation}
\noindent where $E_{1:\mu-1}^j$ denotes a group of $j$-th user's pre-registered events before the event $e_{\mu}$ (i.e., calendar context). In this way, we choose a time slot that maximizes the satisfaction of multiple users. Note that the number of pre-registered events may differ between users. Also, while we have assumed all users have the same influence in multi-attendee event scheduling, more sophisticated aggregation such as multiplying a weighting factor for each user is possible. However, we use the simplest form of aggregation to test the effectiveness of each model trained on personal event scheduling data.

\section{Methodology}
\begin{figure}[t]
\includegraphics[width=8.5cm]{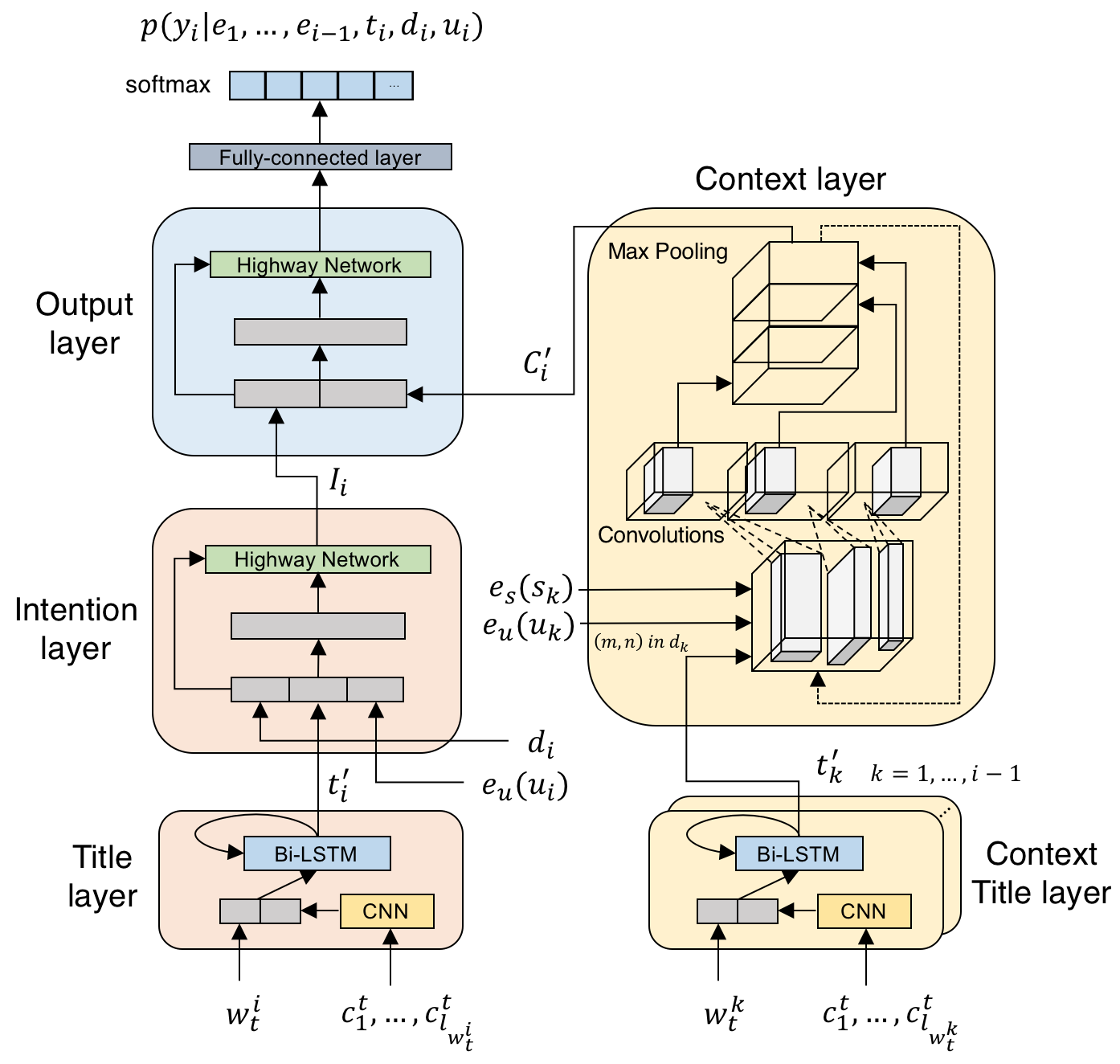}
\caption{NESA overview. Given the title, duration, user attributes, and pre-registered events, NESA suggests suitable time slots for events.}
\label{fig:task_overview}
\end{figure}

To deal with various types of raw calendar attributes, we propose NESA which consists of four different layers: 1) Title layer, 2) Intention layer, 3) Context layer, and 4) Output layer. The Title layer aims to represent the meaning of user written event titles using both the words and characters of the titles. In the Intention layer, our model utilizes title, duration, and user representations to learn user preferences and understand the purpose of events. The Context layer consists of multiple convolutional layers for understanding raw calendar contexts. Finally, the Output layer computes the probability of each time slot based on the representations from each layer. The architecture of NESA is illustrated in Figure \ref{fig:task_overview}.

\subsection{Title Layer}
RNNs have become one of the most common approaches for representing the meaning of written text \cite{mikolov2010recurrent}. Among the various RNN models, state-of-the-art NLP models for question answering \cite{seo2016bidirectional} and named-entity recognition \cite{DBLP:journals/corr/LampleBSKD16} often use not only word-level representations but also character-level representations as inputs. While word-level representations effectively convey semantic/syntactic relationships between words \cite{mikolov2013distributed}, character-level representations are widely used to represent unknown or infrequent words \cite{kim2016character}. In event scheduling tasks, it is essential to use character-level representations for understanding personal calendars that have numerous pronouns or abbreviations.\\ 
\indent Following previous works on question answering, we represent each title $t_i$ using Bi-LSTM \cite{hochreiter1997long,schuster1997bidirectional} with pretrained word embeddings such as GloVe \cite{pennington2014glove}. Given a title $t_i$ comprised of $T_i$ words, we map the words into a set of word embeddings ${w_{1}^i, \dots, w_{T_i}^i}$. The Title layer computes hidden state $h_{T_i}$ of the LSTM as follows:
\begin{equation}
\label{eq:lstm_hidden}
h_{t} = LSTM(w_{t}^i, h_{t-1})
\end{equation}
\noindent where $h_t$ is the $t$-th hidden state of the LSTM which is calculated as follows:
\begin{align*}
i_t &= \sigma (W_{i1}w_t + W_{i2}h_{t-1} + b_{i}) && \\
f_t &= \sigma (W_{f1}w_t + W_{f2}h_{t-1} + b_{f}) && \\
g_t &= \tanh (W_{g1}w_t+ W_{g2}h_{t-1} + b_{g}) &&\\
o_t &= \sigma (W_{o1}w_t + W_{o2}h_{t-1} + b_{o}) && \\
c_t &= f_t \odot c_{t-1} + i_t \odot g_t &&\\
h_t &= o_t \odot \tanh(c_t) &&
\end{align*}
\noindent where we have omitted $i$ from $w_{t}^i$ for clarity and $\odot$ denotes element-wise multiplication. $W_*$ and $b_*$ are trainable parameters of the LSTM. LSTM is effective in representing long-term dependencies between distant inputs using input gate $i$ and forget gate $f$.\\
\indent The Title layer uses Bi-LSTM for title representations. With the forward LSTM giving the final hidden state $h_{T_i}^f$, we build the backward LSTM which computes its hidden states with reversed inputs. The backward LSTM's last hidden state denoted as $h_1^b$ is concatenated with $h_{T_i}^f$ to form the title representation. The title representation will be denoted as $t_i' = [h_{T_i}^f, h_1^b] \in \mathbb{R}^T$.\\
\indent On the other hand, the characters of each word with length $l_{w_t}$ can be represented as a set of character embeddings ${c_{1}^t, \dots, c_{l_{w_t}}^t}$. A common way to combine character embeddings into a word character representation is to use convolutions as follows:
\begin{equation}
\label{eq:char_vec}
f_k^c = \tanh(<C_{k:k+m-1}^t, F> + b)
\end{equation}

\noindent where $f_k^c$ is $k$-th element of a feature map $f^c$, $C^t_{k:k+m-1}$ is a concatenation of character embeddings from $c_k^t$ to $c_{k+m-1}^t$, $m$ is a convolution width, $F$ is a filter matrix, and $<\cdot, \cdot>$ denotes the Frobenius inner product. Using max-over-time pooling \cite{collobert2011natural}, the single scalar feature is extracted as $f^c = \max_k f_k^c$. Given $N$ types of filters and each of them having a different number of filters, resulting word character representations are obtained as $w_t^{c, i} = [f^{c, 1}, \dots, f^{c,N}]$ where $[\cdot, \cdot]$ denotes a vector concatenation, and $f^{c,n}$ is a concatenation of the outputs of $n$-th filters. We concatenate word representation $w_t^i$ with word character representation $w_t^{c,i}$, which is inputted into the LSTM in Equation \ref{eq:lstm_hidden}.
\subsection{Intention Layer}
Users have different intentions when registering a specific event. For instance, event titles that contain only personal names connote \textit{meetings} to someone, but could mean \textit{appointments} to others. To capture the intention of each user, we incorporate the title $t_i$, duration $d_i$, and user $u_i$ attributes in the Intention layer. In this way, the Intention layer takes into account user preferences and purposes of events. In particular, we use the Highway Network that has some skip-connections between layers \cite{srivastava2015highway}.\footnote{While we could use Multi-Layer Perceptron (MLP) instead, the Highway Network achieved better performance in our preliminary experiments.} Given a title representation $t_i'$ from a Title layer, duration $d_i$, and user $u_i$, the output of the Highway Network $I_i$ is as follows:
\begin{equation}
\label{eq:highway_x}
x = [t_i', d_{i}, e_u(u_{i})]
\end{equation}
\begin{equation}
\label{eq:highway_q}
q = \sigma (W_qx + b_q)
\end{equation}
\begin{equation}
\label{eq:highway_ii}
I_i = q \odot f(W_hx + b_h) + (1-q) \odot x
\end{equation}
\noindent where $e_u(\cdot) \in \mathbb{R}^U$ is an embedding mapping for each user. $W_q\textbf{, }W_h \in \mathbb{R}^{(T+U+1) \times (T+U+1)}$ are trainable parameters and $f$ is a nonlinearity. Due to the skip-connection from $x$ to $I_i$ in addition to the nonlinearity, the Intention layer easily learns both linear and nonlinear relationships between calendar attributes.
% * <susanniekim@gmail.com> 2018-05-19T16:33:04.260Z:
% 
% > , 
% and?
% 
% ^ <krth32@gmail.com> 2018-05-20T13:50:52.897Z:
% 
% as a mathematical expression, that expression (,) seems okay
% 
% ^ <susanniekim@gmail.com> 2018-05-21T17:31:49.762Z.

\subsection{Context Layer}
We define a calendar context as a set of events that are pre-registered before the event $e_i$. We denote each pre-registered event as $e_k$ where $k$ is from $1$ to $i-1$. Note that each user's week has a varying number of events from 0 to more than 50. Also, each pre-registered event $e_k$ is comprised of different kinds of attributes such as start time, title, and duration. In the Context layer, we represent the calendar context by reflecting the status of the current week and scheduling preferences of users. Then, we use CNN to capture the local and global features of the calendar context to understand the calendar context representation.
% * <susanniekim@gmail.com> 2018-05-19T16:42:49.516Z:
% 
% > we use CNN to understand the calendar context representation by capturing the local and global features of the calendar context.
% we use CNN to capture the local and global features of the calendar context to understand the calendar context representation.
% 
% ^ <krth32@gmail.com> 2018-05-20T09:42:53.984Z:
% 
% modified to your sentence
%
% ^ <susanniekim@gmail.com> 2018-05-21T17:34:52.019Z.
% * <susanniekim@gmail.com> 2018-05-19T16:37:20.382Z:
% 
% > preference
% s?
% 
% ^ <krth32@gmail.com> 2018-05-20T09:40:41.610Z:
% 
% added s
%
% ^ <susanniekim@gmail.com> 2018-05-21T17:35:24.914Z.

\subsubsection{Context Title Representation}
For each title $t_k$ in a pre-regist\-er\-ed event $e_k$, we build a Context Title layer that processes only the titles of pre-registered events. Using Bi-LSTM and character-level CNN, each context title representation is obtained as $t_k'$. Note that multiple context title representations are obtained simultaneously in a mini-batch manner.

\subsubsection{Calendar Context Representation}
Given the context title representations $t'_k \in \mathbb{R}^T$, we construct a calendar context $C_i \in \mathbb{R}^{(M \times N) \times (T+U+S)}$ where $U$ and $S$ are dimensions of user and slot embeddings, respectively. $M$ represents the number of days in a week, and $N$ represents the number of hours in a day. Each depth is denoted as $C_i^{m, n} \in \mathbb{R}^{T+U+S}$ which is from $m$-th row (day) and $n$-th column (hour) of $C_i$. Each $C_i^{m,n}$ is constructed as follows:
\begin{equation}
\label{eq:cal_context_c}
C_i^{m,n} = [t'_{(m,n)}, e_u(u_i), e_s(s_{(m,n)})]
\end{equation}
\begin{equation}
\label{eq:cal_context_tprime}
t'_{(m,n)} = \left\{
	\begin{array}{ll}
    t'_k & \text{if } (m, n)\text{ lies in }e_k\text{'s duration } d_k \\
    \textbf{0} \in \mathbb{R}^T & \text{otherwise}
     \end{array}\right\}
\end{equation}

\noindent where $e_u(\cdot) \in \mathbb{R}^U$ and $e_s(\cdot) \in \mathbb{R}^S$ are user and slot embedding functions, respectively, and $s_{(m,n)}$ is a slot representation on $m$-th day at $n$-th hour.\\
% * <susanniekim@gmail.com> 2018-05-19T16:49:32.129Z:
% 
% > in
% on ..day and at ... hour?
% 
% ^ <krth32@gmail.com> 2018-05-20T09:44:38.661Z:
% 
% modified it like this
%
% ^ <susanniekim@gmail.com> 2018-05-21T17:36:17.513Z.
\indent Given the calendar context $C_i$, the first convolution layer convolves $C_i$ with 100 ($1\times 1$), 200 ($3\times 3$), 300 ($5\times 5$) filters, followed by batch normalization \cite{Ioffe:2015:BNA:3045118.3045167} and element-wise rectifier nonlinearity. We pad the calendar context to obtain same size outputs for each filter, and concatenate each output depth-wise. The second convolution layer consists of 50 ($1\times 1$), 100 ($3\times 3$), 150 ($5\times 5$) filters, followed by batch normalization and max-over-time pooling. As a result, we obtain the final calendar context representation $C'_i \in \mathbb{R}^{300}$.
% * <susanniekim@gmail.com> 2018-05-19T16:56:56.476Z:
% 
% > pad
% uc 
% 
% ^ <krth32@gmail.com> 2018-05-20T09:45:34.917Z:
% 
% it is a academic term
%
% ^ <susanniekim@gmail.com> 2018-05-21T17:33:56.487Z.

\subsection{Output Layer}
Given a calendar context representation $C'_i$ and an intention representation $I_i$, the Output layer computes the probability of each time slot in $M \times N$. We again adopt a Highway Network to incorporate the calendar context representation and the intention representation. Similar to Equations \ref{eq:highway_x}-\ref{eq:highway_ii}, given the input $x_o = [C'_i, I_i]$, the probability distribution of time slots is as follows:
% * <susanniekim@gmail.com> 2018-05-19T17:04:34.209Z:
% 
% > to
% and? or Equations 5-7? 
% 
% ^ <krth32@gmail.com> 2018-05-20T09:48:09.832Z:
% 
% Equations 5-7 is close,  but conferences recommend to use full names. how about "from Equation (5) to Equation (7)"?
% 
% ^ <susanniekim@gmail.com> 2018-05-21T17:39:37.564Z:
% 
% Similar to from is uc 
% Equation 5, eq 6, and eq 7
%
% ^ <dweller92@naver.com> 2018-05-22T14:19:08.072Z:
% 
% eq 5-7 seems fine
%
% ^ <dweller92@naver.com> 2018-05-22T14:44:06.663Z.
\begin{equation}
\label{eq:output_z}
z = q_o \odot f(W_hx_o + b_h) + (1-q_o) \odot x_o
\end{equation}
\begin{equation}
\label{eq:output_p}
p_j = softmax(W_{o} z + b_{o})_j
\end{equation}
\begin{equation}
\label{eq:softmax}
softmax(\boldsymbol{\alpha})_j = \frac{exp(\alpha_j)} {\sum_{j^\prime}exp(\alpha_{j^\prime})}
\end{equation}
\noindent where $q_o$ is obtained in the same way as Equation \ref{eq:highway_q} and $j$ is from 1 to $M \times N$. We have used a single fully-connected layer for predicting the start time slot $y_i$ of the event $e_i$. Given the outputs, the cross-entropy loss $CE(\Theta)$ of NESA is computed as follows:
\begin{equation}
\label{eq:output_loss}
CE(\Theta) = -\dfrac{1}{K}\sum_{i=1}^{K} \log p(y_i|e_1, \dots, e_{i-1}, t_i, d_i, u_i;\Theta)
\end{equation}
\noindent where $K$ denotes the number of events in a week. The model is optimized on the weeks in the training set. We use the Adam optimizer \cite{kingma2014adam} to minimize Equation \ref{eq:output_loss}.

\begin{table}
\caption{Event Scheduling Dataset Statistics}
\label{table:data_statistics}
\begin{tabular}{lrr}
\toprule
Statistics & Personal & Multi-Attendee \\
\midrule
\# of users & 859 & 260 \\
\# of unseen users\footnotemark & -- & 217 \\
\# of events & 593,207 & 1,354 \\
\# of weeks & 109,843 & 1,045 \\
Avg. \# of pre-registered events & 6.9 & 22.2 \\
Avg. \# of attendees & -- & 2.1 \\
\bottomrule
\end{tabular}
\end{table}
\footnotetext{The number of users not seen in the personal event scheduling dataset.}

\section{Experiment}
\subsection{Dataset}
\subsubsection{Preprocessing}
We used Google Calendar\footnote{https://www.google.com/calendar} data collected between April 2015 and March 2018 by Konolabs, Inc. The format of the data is based on iCalendar, which is the most commonly used online calendar format. We detected and removed noisy events from the raw calendar data to reflect only real online calendar events. Events that we considered as noise are as follows:
% * <krth32@gmail.com> 2018-05-22T10:54:13.362Z:
% 
% > data
% it was "dataset"
% 
% ^ <dweller92@naver.com> 2018-05-22T14:44:59.887Z.
\begin{itemize}
\item Events automatically generated by other applications (e.g., phone call logs, weather information, and body weight).
%\item Group calendars shared and used by more than 2 users.
\item Having an event title that has no meaning (e.g., empty string).
\item All-day events, i.e., the events that will be held all day long.
\end{itemize}
% Although there are users of various nationalities, we selected only English-speaking users for event scheduling to utilize the pretrained word embeddings in our model. 
\indent Although some of the all-day events are legitimate events such as vacations or long-term projects, most of them are regular events whose start times have been simply omitted by users. We represented time slots as integers ranging from 0 to 167 where each time slot was considered as one hour in a week (i.e., 7 days $\times$ 24 hours). Only one event was selected given the overlapping events. The duration of each event is scaled to a scalar value from 0 to 1.\\
% * <susanniekim@gmail.com> 2018-05-19T17:33:16.406Z:
% 
% > rescaled
% converted?
% 
% ^ <krth32@gmail.com> 2018-05-20T09:58:23.292Z:
% 
% academic term. how about "normalized"?
%
% ^ <susanniekim@gmail.com> 2018-05-21T17:41:57.206Z:
% 
% it's your decision. 
%
% ^ <krth32@gmail.com> 2018-05-22T11:27:30.538Z:
% 
% I will keep "rescaled"
%
% ^ <krth32@gmail.com> 2018-05-22T11:27:36.259Z.
\indent In Table \ref{table:data_statistics}, the second column shows the statistics of the personal event scheduling dataset after filtering. Though we carefully filtered calendar events, the dataset still had a considerable number of unrecognizable events (e.g., personal abbreviations). However, to test the ability of our proposed model, we did not perform any further filtering. We split the dataset into training (80\%), validation (10\%), and test (10\%) sets, respectively. \\ %Due to the privacy issues of the data, we offer a test platform on which users can preprocess their own calendar data. \\
% * <krth32@gmail.com> 2018-05-22T10:53:13.649Z:
% 
% > events
% it was "users"
% 
% ^ <dweller92@naver.com> 2018-05-22T14:45:22.698Z.
\indent In Table \ref{table:data_statistics}, the third column shows the statistics of the multi-attendee event scheduling dataset. Each event in the multi-attendee event scheduling dataset has at least two attendees, and attendees in each event are in the same time zone.\footnote{This can be easily extended to different time zone situations by shifting one of the time offsets.} Due to the small number of multi-attendee events, we use them only as a test set for multi-attendee event scheduling. Also, we ensure that no events in the multi-attendee event scheduling dataset appear in the personal event scheduling dataset. As the multi-attendee event scheduling dataset has multiple attending users per event, it has more pre-registered events (22.2) than the personal event scheduling dataset (6.9). Note that both the personal and multi-attendee event scheduling datasets have a much larger number of users than the CAP dataset\footnote{The CAP dataset contains system logs of Calendar Apprentice, which are difficult to convert to the iCalendar format.} which has events of only 2 users \cite{mitchell1994experience,zunino2009chronos}.

\subsubsection{Evaluation Metrics}
We used various metrics to evaluate the performance of each model in event scheduling. Recall@N is the metric that determines if the correct time slot is in the top $n$ predictions. Recall@1 and Recall@5 were mainly used. We also used Mean Reciprocal Rank (MRR) which is the mean of the inverse of the correct answer's rank. Also, motivated by the fact that suggesting time slots close to the correct answers counts as proper event scheduling, we used Inverse Euclidean distance (IEuc) \cite{toby2007programming} which calculates the inverse distance between predicted slots $\hat{y}_i$ and answer slots $y_i$ in two-dimensional space in terms of days (m) and hours (n) as follows: 
% * <krth32@gmail.com> 2018-05-22T14:03:33.463Z:
% 
% > (n)
% added
% 
% 
% ^ <dweller92@naver.com> 2018-05-22T14:45:51.115Z.
% * <krth32@gmail.com> 2018-05-22T14:03:25.094Z:
% 
% > (m)
% added
% 
% 
% ^ <dweller92@naver.com> 2018-05-22T14:45:52.209Z.
\begin{equation}
\label{eq:euc}
Euc(\hat{y}_i, y_i) = \sqrt{(\hat{y}_{i_{m}}-y_{i_{m}})^2+(\hat{y}_{i_{n}}-y_{i_{n}})^2}
\end{equation}
\begin{equation}
\label{eq:ieuc}
IEuc(\hat{y}_i, y_i) = \dfrac{1}{Euc(\hat{y}_i, y_i) + 1}.
\end{equation}

\begin{table}
\small
\caption{Hyperparameters of MLP and NESA}
\label{table:model_parameter}
\begin{tabular}{rlr}
\toprule
Model & Parameter & Value \\
\midrule
MLP & Hidden layer size & 500 \\
& \# of hidden layers & 2 \\
& Learning rate & 0.0005 \\
\midrule
NESA & LSTM cell hidden size & 100 \\
& \# of LSTM layers & 2 \\
& LSTM dropout & 0.5 \\
& Day $M$, hour $N$ & 7, 24 \\
& $T$, $C$, $S$, $U$ & 200, 30, 30, 30 \\
% & Kernel sizes (w, h) & [(1, 1), (3, 3), (5, 5)] \\
% & \# of total filters ($K_l$) & [600, 300] \\
& Learning rate & 0.001 \\
\bottomrule
\end{tabular}
\end{table}

\subsection{Experimental Settings}
\subsubsection{Baseline Models}
\indent While recent automatic scheduling systems have proven to be effective on small sized datasets \cite{wainer2007scheduling,zunino2009chronos,cranshaw2017calendar}, it is difficult to directly apply their methodologies to our tasks for the following reasons: 1) some of them assume that user preferences are already given \cite{wainer2007scheduling}, 2) some use learning mechanisms based on systematic interactions with users \cite{zunino2009chronos}, or 3) require human labor \cite{cranshaw2017calendar}. As a result, we use baseline models that are easily reproducible but still effective in our tasks.\\
% * <susanniekim@gmail.com> 2018-05-19T18:31:06.565Z:
% 
% > focus
% use?
% 
% ^ <krth32@gmail.com> 2018-05-20T10:10:20.009Z:
% 
% I think "focus" is right here
%
% ^ <susanniekim@gmail.com> 2018-05-21T17:50:42.312Z:
% 
% focus is a bit unclear. what do you mean by focus? you study them? 
%
% ^ <krth32@gmail.com> 2018-05-22T14:34:30.064Z:
% 
% my thoughts were unclear. modified to "use"
%
% ^ <dweller92@naver.com> 2018-05-22T14:46:34.724Z.
\indent In our study, the baselines are as follows: 1) a variant of CAP \cite{mitchell1994experience} using Random Forest (RF), 2) Support Vector Machine (SVM) \cite{gervasio2005active,berry2011ptime}, 3) Logistic Regression (LogReg), and 4) Multi-Layer Perceptron (MLP). While RF and SVM are representative of previously suggested scheduling models, we further evaluate LogReg and MLP which are frequently adopted as classification baseline models.\\
% * <susanniekim@gmail.com> 2018-05-22T19:05:17.630Z:
% 
% > Logistic Regression (LR), and 4) Multi-Layer Perceptron (MLP)
% LR model, MLP model? 
% 
% ^ <dweller92@naver.com> 2018-05-23T06:33:25.703Z:
% 
% just pronouns.
%
% ^ <dweller92@naver.com> 2018-05-23T06:34:09.700Z.
% * <susanniekim@gmail.com> 2018-05-22T19:03:39.984Z:
% 
% > Support Vector Machine (SVM) 
% model?
% 
% ^ <dweller92@naver.com> 2018-05-23T06:33:46.530Z:
% 
% no
%
% ^ <dweller92@naver.com> 2018-05-23T06:34:11.182Z.
% * <susanniekim@gmail.com> 2018-05-19T18:40:37.050Z:
% 
% > . While RF and SVM represent previously suggested scheduling models
% uc since they...? represent or used as sched models?
% 
% ^ <krth32@gmail.com> 2018-05-20T10:13:02.724Z:
% 
% how about "While RF and SVM are representative of previously suggested scheduling models"?
%
% ^ <susanniekim@gmail.com> 2018-05-21T17:52:47.059Z.
% * <susanniekim@gmail.com> 2018-05-19T18:39:23.582Z:
% 
% > Considered baseline models are 
% We used the following base models? 
% 
% ^ <krth32@gmail.com> 2018-05-20T10:17:12.902Z:
% 
% we didn't. 
% previous studies used RF related model and  SVM
% how about "Considered as baseline models are"?
% 
% ^ <susanniekim@gmail.com> 2018-05-21T17:56:36.019Z:
% 
% Prev studies have used the following baseline models:? 
% 
% ^ <krth32@gmail.com> 2018-05-22T15:26:16.822Z:
% 
% Prev studies have used RF and SVM. 
% would "In our study, baseline models are" be better? 
%
% ^ <susanniekim@gmail.com> 2018-05-22T19:06:52.014Z:
% 
% check again please
%
% ^ <dweller92@naver.com> 2018-05-23T06:33:58.170Z:
% 
% good.
%
% ^ <dweller92@naver.com> 2018-05-23T06:34:03.786Z.
\indent As previous studies have focused on building interactive scheduling software, their learning algorithms rely largely on system dependent features such as event types, position of attendees, names of college classes, and so on \cite{mitchell1994experience}. As the iCalendar format does not contain most of these system dependent features, we used the attributes in Section 3.1 as inputs to the four baseline models. Besides categorical or real-valued features, event titles are represented as the average of pretrained word embeddings, and calendar contexts are given as binary vectors in which filled time slots are indicated as 1. For user representations, we used the normalized event start time statistics of each user (i.e., 168 dimensional vector whose elements sum to 1.) to reflect the scheduling preferences of each user. The representation of an unseen user is obtained using the average start time statistics of all the users in the training set.\footnote{Each baseline feature representation was selected among various hand-crafted features based on our in-house experiments. For instance, statistics based user representation was better than one-hot user representation in terms of both event scheduling performance and generalization.} The biggest difference between the baseline models and NESA is that the baseline models use a fixed set of hand-crafted features, whereas NESA learns to represent user preferences and calendar contexts for effective event scheduling.

\subsubsection{Model Settings}
\indent While CAP uses a single decision tree for event scheduling, we constructed RF using thousand decision trees to build a more effective baseline model. The SVM model uses squared hinge losses and the one-vs-rest strategy for training. For LogReg, we used the SAGA optimizer \cite{defazio2014saga}. Rectified linear unit (ReLU) \cite{nair2010rectified} was used for MLP's activation function. Also for MLP, early stopping was applied based on the loss on the validation set, and we used the Adam optimizer for MLP. Both LogReg and MLP used $L_2$ regularizations to avoid overfitting.\\
% * <susanniekim@gmail.com> 2018-05-21T18:14:08.647Z:
% 
% > RF to a more effective baseline} model.
% to build a more effective baseline RF model?
% 
% ^ <krth32@gmail.com> 2018-05-22T11:44:54.640Z:
% 
% I changed that as you said.
%
% ^ <krth32@gmail.com> 2018-05-22T14:06:10.608Z.
\indent The hyperparameters of MLP and NESA were chosen based on the MRR scores on the validation sets and the results are shown in Table \ref{table:model_parameter}. We used the same hyperparameters from \cite{kim2016character} for character-level convolutions. A dropout of 0.5 was applied to the non-recurrent part of the RNNs of NESA to prevent overfitting \cite{zaremba2014recurrent}. We also clipped gradients when their norm exceeded 5 to avoid exploding gradients. Besides the character embedding, there are three additional embeddings in NESA: 1) word, 2) user, and 3) slot. We used pretrained GloVe\footnote{For both NESA and baseline features, we used glove.840B.300d word embeddings.} for word embeddings, and randomly initialized embeddings for character, user, and slot embeddings. Word embeddings were fixed during optimization while other embeddings were optimized during training.\\
% * <susanniekim@gmail.com> 2018-05-19T19:36:01.018Z:
% 
% > the RNN
% whose RNN?
% 
% ^ <krth32@gmail.com> 2018-05-20T14:07:41.026Z:
% 
% modified to "the RNNs of NESA"
% 
% ^ <susanniekim@gmail.com> 2018-05-21T17:59:51.999Z.
% * <susanniekim@gmail.com> 2018-05-19T19:34:20.767Z:
% 
% > results
% same as MRR scores?
% 
% ^ <krth32@gmail.com> 2018-05-20T14:12:22.563Z:
% 
% related, but different.
% modified to "optimization results"
%
% ^ <susanniekim@gmail.com> 2018-05-21T18:00:56.163Z.
\indent For training NESA, we used PyTorch with a CUDA enabled NVIDIA TITAN Xp GPU. The baseline models were trained using Scikit-learn. It took 8 hours of training for NESA to converge, which is quite short given the size of our training set and the complexity of NESA. NESA performs event scheduling as fast as baseline models by using mini-batches. We also experimented with increased number of layers and hidden dimensions in the MLP model so that it would have the same number of parameters as NESA (8.5M). However, the performance of the MLP model was lower than that of the MLP model trained on the best hyperparameters (7.0\% in terms of MRR).

\begin{table}
\caption{Personal Event Scheduling Results}
\label{table:personal_result}
\begin{tabular}{lrrrrrr}
\toprule
Model & Recall@1 & Recall@5 & MRR & IEuc \\
\midrule
RF \cite{mitchell1994experience} & 0.0348 & 0.1483 & 0.0988 & 0.2520 \\
SVM \cite{gervasio2005active,berry2011ptime} & 0.0445 & 0.1762 & 0.1271 & 0.2619 \\
LogReg & 0.0442 & 0.1749 & 0.1279 & 0.2678 \\
MLP & 0.0442 & 0.1803 & 0.1277 & 0.2725 \\
NESA & \textbf{0.0604} & \textbf{0.2156} & \textbf{0.1542} & \textbf{0.2881} \\
\bottomrule
\end{tabular}
\end{table}

\begin{table}
\caption{Multiple Attendee Event Scheduling Results}
\label{table:multi_result}
\begin{tabular}{lrrrrrr}
\toprule
Model & Recall@1 & Recall@5 & MRR & IEuc \\
\midrule
RF \cite{mitchell1994experience} & 0.0635 & 0.2585 & 0.0742 & 0.2389 \\
SVM \cite{gervasio2005active,berry2011ptime} & 0.0030 & 0.0340 & 0.0234 & 0.2530 \\
LogReg & 0.0037 & 0.0332 & 0.0260 & 0.2608 \\
MLP & 0.0406 & 0.1928 & 0.0773 & 0.2507 \\
NESA & \textbf{0.0960} & \textbf{0.2740} & \textbf{0.1744} & \textbf{0.2950} \\
\bottomrule
\end{tabular}
\end{table}

\subsection{Quantitative Analysis}
\subsubsection{Personal Event Scheduling}
The scores of personal event scheduling are presented in Table \ref{table:personal_result}. The reported scores are average test set scores after ten separate trainings. The best scores are in bold. We first see that the performance ranking of the IEuc scores is similar to that of other metric scores such as the Recall@5 scores. This shows that the more a model accurately predicts an answer, the more it suggests nearby time slots around the correct answer. Among the baseline models, MLP performed the best on average, and RF achieved the lowest overall scores. However, despite MLP's deeper structure, performance improvements of MLP over LogReg were marginal, which shows the limitation of feature based models. NESA achieved higher scores than the baseline models in all metrics by learning to schedule directly using raw calendar data. NESA outperformed the baseline models by 29.6\% on average in terms of MRR. More specifically, NESA outperformed MLP, which is the best baseline model, by 36.5\%, 19.6\%, 20.7\%, and 5.7\% in terms of Recall@1, Recall@5, MRR, and IEuc, respectively.

\subsubsection{Multi-Attendee Event Scheduling}
The performance results of the models on multi-attendee event scheduling are presented in Table \ref{table:multi_result}. The scores of each model are obtained by Equation \ref{eq:multi}. Compared to the performances on personal event scheduling, Recall@1 and Recall@5 of RF have been greatly improved, but MRR and IEuc of RF have been degraded. This verifies the limited effectiveness of decision tree based models as reported in the work of Mitchell et al. \cite{mitchell1994experience}. RF fails to provide precise probability distribution over time intervals, that reflects user preferences and calendar contexts, as MRR and IEuc are more sensitive to suggestion quality over the whole week. Other baseline models such as SVM, LogReg, and MLP have failed to produce meaningful results on multi-attendee event scheduling. We found that the huge performance degradation of these models comes from generalization failure on unseen users as most users (217 out of 260) in the multi-attendee event scheduling dataset are unseen during training on the personal event scheduling dataset. The performance of SVM, LogReg, and MLP on multi-attendee event scheduling was higher (but still insufficient compared to RF and NESA) when all the attendees were comprised of seen users during training.\\
\indent NESA does not suffer from the unseen user problem by understanding raw online calendars to infer user preferences and understand calendar contexts. While preferences of known users can be encoded in user embeddings in NESA, preferences of unseen users can be inferred from their raw calendars. As with the personal event scheduling task, NESA outperforms the other baseline models by large margins on the multi-attendee event scheduling task. Specifically, NESA outperforms the best baseline model RF by 51.2\%, 6.0\%, 135.0\%, and 23.5\% in terms of Recall@1, Recall@5, MRR, and IEuc, respectively. This shows that using raw calendar data for understanding user preferences and calendar contexts is very important in event scheduling tasks.
% * <susanniekim@gmail.com> 2018-05-21T19:11:50.795Z:
% 
% > raw user calendars
% raw cale data?
% 
% ^ <krth32@gmail.com> 2018-05-22T11:58:31.591Z:
% 
% modified to "raw calendar data"
%
% ^ <krth32@gmail.com> 2018-05-22T14:15:23.781Z.
% * <susanniekim@gmail.com> 2018-05-19T20:28:13.901Z:
% 
% >  results of multi-attendee event schedulin
% perfor of the  models on the ....scheduling?
% 
% ^ <krth32@gmail.com> 2018-05-20T14:27:24.145Z:
% 
% modified to "performance of the models on the multi-attendee event scheduling"
%
% ^ <susanniekim@gmail.com> 2018-05-21T19:08:06.794Z.
% * <susanniekim@gmail.com> 2018-05-19T20:27:51.050Z:
% 
% > with pretrained models
% refers to what?
% 
% ^ <krth32@gmail.com> 2018-05-20T14:28:21.795Z:
% 
% modified to "with pretrained RF and NESA models"
%
% ^ <susanniekim@gmail.com> 2018-05-21T19:20:41.477Z:
% 
% EQ has RF and nesa models? uc  
%
% ^ <dweller92@naver.com> 2018-05-22T14:52:38.056Z:
% 
% all models including baselines and NESA.
% not necessary so I deleted it.
% 
% ^ <dweller92@naver.com> 2018-05-22T14:53:59.344Z.
% * <susanniekim@gmail.com> 2018-05-19T20:26:48.602Z:
% 
% > training
% tests?
% 
% ^ <krth32@gmail.com> 2018-05-20T14:28:57.376Z:
% 
% training is right here
%
% ^ <susanniekim@gmail.com> 2018-05-21T19:09:18.035Z.

\begin{table}
\small
\captionsetup{justification=centering}
\caption{NESA Model Ablation \\ (Diff. \%: average performance difference \% of 4 metrics)}
\label{table:nesa_ablation}
\begin{tabular}{lrrrrrrr}
\toprule
Model & Recall@1 & Recall@5 & MRR & IEuc & Diff. \%\\
\midrule
NESA & 0.0623 & 0.2289 & 0.1605 & 0.2910 & -- \\
- Context L. & \underline{0.0419} & 0.1789 & \underline{0.1083} & 0.2668 & \underline{-23.9} \\
- Intention L. & 0.0444 & \underline{0.1657} & 0.1234 & \underline{0.2614} & -22.4 \\
- Word E. & 0.0561 & 0.2079 & 0.1476 & 0.2783 & -7.9 \\
- Character E. & 0.0518 & 0.1974 & 0.1418 & 0.2836 & -11.2 \\
- Duration F. & 0.0572 & 0.2049 & 0.1477 & 0.2820 & -7.4 \\
- User E. & 0.0587 & 0.2125 & 0.1522 & 0.2889 & -4.7 \\
\bottomrule
\end{tabular}
\end{table}

\begin{table}
\small
% \captionsetup{justification=centering}
\caption{Baseline Model Ablation}
% * <susanniekim@gmail.com> 2018-05-19T20:55:49.440Z:
% 
% > Diff. \%
% what does this mean? 
% 
% ^ <krth32@gmail.com> 2018-05-20T14:30:11.451Z:
% 
% that means an average percentage of performance differences
% 
% ^ <susanniekim@gmail.com> 2018-05-21T19:19:02.681Z:
% 
% would avg. perf diff %  be better?
%
% ^ <krth32@gmail.com> 2018-05-22T12:03:15.265Z:
% 
% modified the description to "average performance difference %"
%
% ^ <krth32@gmail.com> 2018-05-22T15:10:21.703Z.
\label{table:baseline_ablation}
\begin{tabular}{lrrrrrrr}
\toprule
Model & Recall@1 & Recall@5 & MRR & IEuc & Diff. \% \\
\midrule
MLP & 0.0445 & 0.1805 & 0.1283 & 0.2719 & -- \\
- Context F. & \underline{0.0384} & \underline{0.1624} & \underline{0.1026} & \underline{0.2582} & \underline{-12.2} \\
- Word F. & 0.0425 & 0.1710 & 0.1245 & 0.2661 & -3.7 \\
- Character F. & 0.0433 & 0.1788 & 0.1271 & 0.2724 & -1.1 \\
- Duration F. & 0.0433 & 0.1760 & 0.1256 & 0.2704 & -2.0 \\
- User F. & 0.0440 & 0.1790 & 0.1269 & 0.2722 & -0.7 \\
\bottomrule
\end{tabular}
\end{table}

\begin{figure}
\hspace*{-0.1in}
\includegraphics[width=8.2cm]{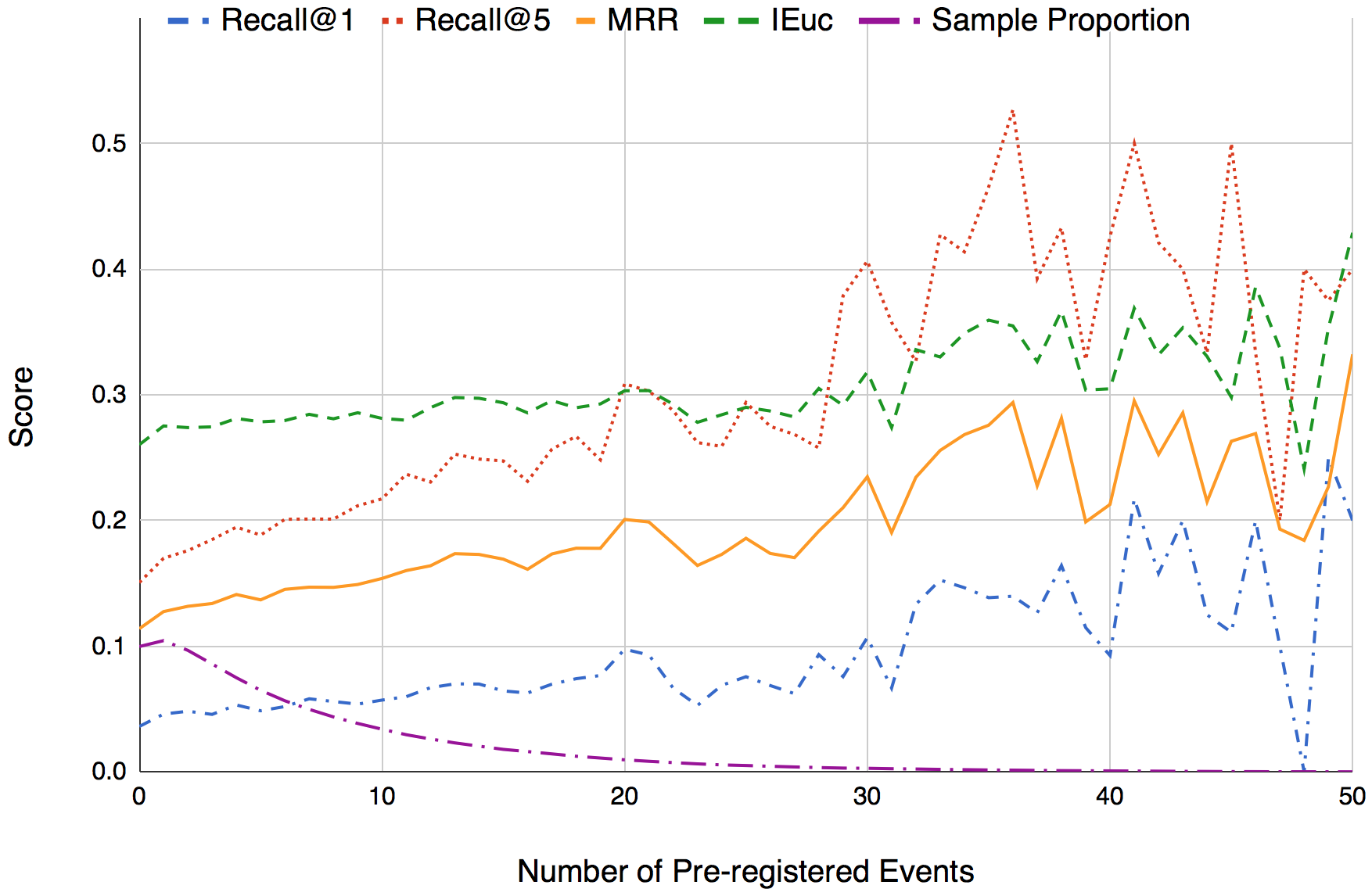}
\caption{Performance \textbf{changes with different numbers} of pre-registered events in NESA.}
% * <susanniekim@gmail.com> 2018-05-19T20:58:11.317Z:
% 
% > changes with different numbers
% changes due to the diff? 
% 
% ^.
\label{fig:perform_context_num}
\end{figure}

% \begin{figure}
% \includegraphics[width=8.4cm]{figures/perform_multi_attendee.png}
% \caption{Performance change with different number of attendees in NESA.}
% \label{fig:perform_attendee_num}
% \end{figure}

\begin{figure}[t]
\includegraphics[width=8.5cm]{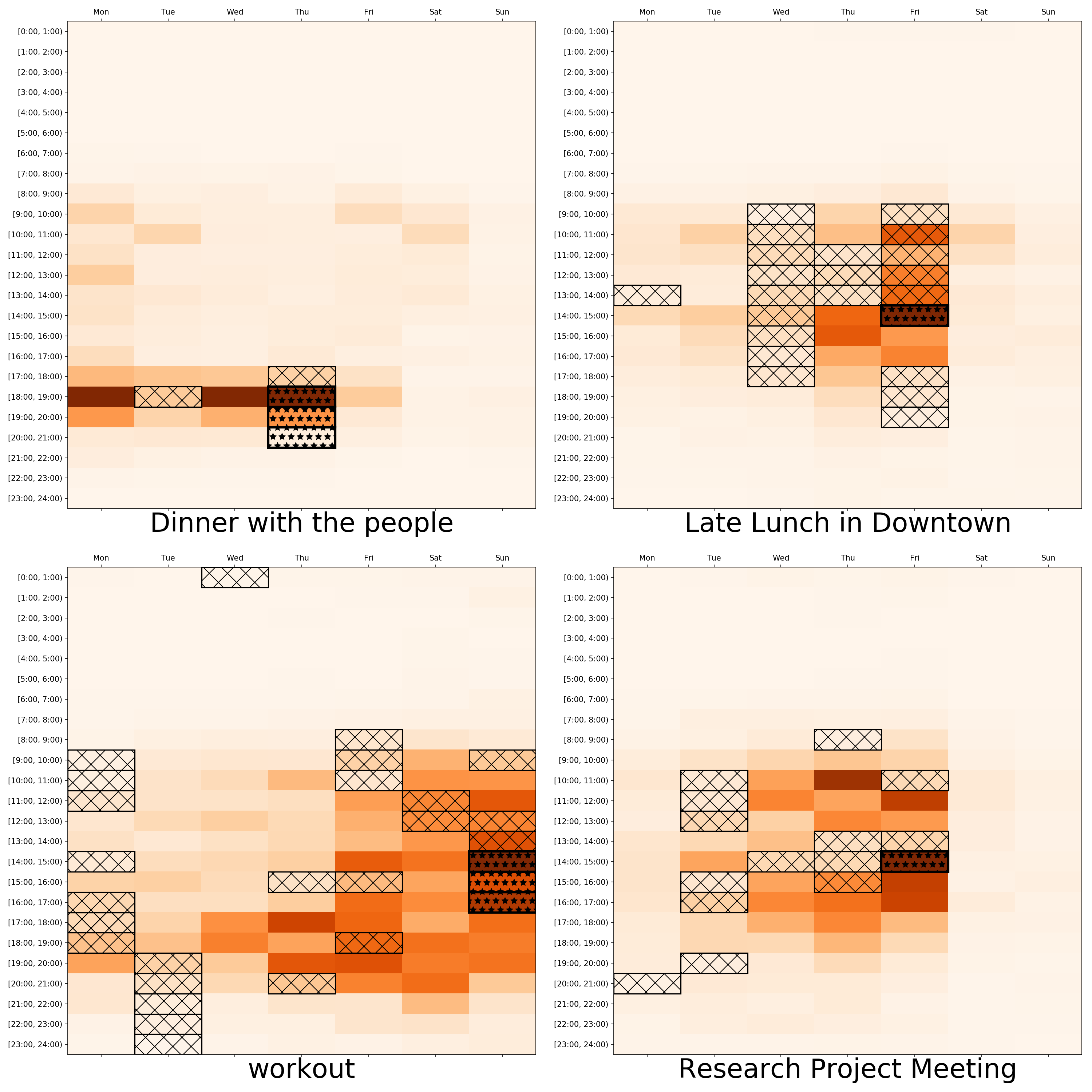}
\caption{Output probabilities of NESA given different titles.}
\label{fig:title_heatmap}
\end{figure}

\subsubsection{NESA Model Ablation and Analysis}
To analyze the architecture of NESA, we removed each layer or component of NESA. The results are shown in Table \ref{table:nesa_ablation}. When the Context layer is removed, the Output layer receives only the intention representation. We feed the title representation instead of the intention representation to the Output layer when the Intention layer is removed. The Context layer has the most significant impact on the overall performance. The Intention layer also shows that incorporating user and duration attributes with title attributes is crucial for event scheduling. The character embedding has substantial effects on the performance.\\ 
% * <susanniekim@gmail.com> 2018-05-22T20:05:48.075Z:
% 
% >  character embedding 
% uc because you don't mention it in this section
% 
% ^ <dweller92@naver.com> 2018-05-23T06:41:43.217Z.
% * <susanniekim@gmail.com> 2018-05-21T19:34:27.429Z:
% 
% > features
% or attri? in previous sentence you state dur att
% 
% ^ <krth32@gmail.com> 2018-05-22T12:04:01.334Z:
% 
% modified to "attributes"
%
% ^ <krth32@gmail.com> 2018-05-22T12:04:12.671Z.
% * <susanniekim@gmail.com> 2018-05-19T20:47:00.028Z:
% 
% > feed
% uc real term? feed to what? 
% 
% ^ <krth32@gmail.com> 2018-05-20T14:40:01.704Z:
% 
% an acedemic expression
% feed to the Output layer
% 
% ^ <susanniekim@gmail.com> 2018-05-21T19:25:07.183Z.
% * <susanniekim@gmail.com> 2018-05-19T20:42:35.145Z:
% 
% >  intention representation
% real term? the rep of an intent
% 
% ^ <krth32@gmail.com> 2018-05-20T14:41:45.945Z:
% 
% intention representation is an expression in this study
%
% ^ <susanniekim@gmail.com> 2018-05-21T19:30:32.550Z.
\indent To demonstrate the Context layer's impact, we illustrate the changes in performance of NESA based on different numbers of pre-registered events in Figure \ref{fig:perform_context_num}. As the number of pre-registered events grows, overall performance improves. Note that the sampling proportion decreases as the number of pre-registered events increases, which causes a high variance in performance.

\begin{table}
\caption{Nearest Neighbors (NNs) of Title Representations Given the Title \texttt{Family lunch}}
% * <susanniekim@gmail.com> 2018-05-22T20:41:11.910Z:
% 
% > NN
% NNs?
% 
% ^ <dweller92@naver.com> 2018-05-23T06:41:55.183Z:
% 
% yes
%
% ^ <dweller92@naver.com> 2018-05-23T06:41:57.214Z.
\label{table:effect_title}
\begin{tabular}{cc}
\toprule
MLP NNs & NESA NNs \\
\midrule
Family Dinner out & Birthday \underline{lunch} \\
Family Dinner & Themed \underline{Lunch} \\
\underline{Lunch} with Family Friends & UNK / BDP \underline{lunch} \\
family dinner & Hope \underline{lunch} \\
\bottomrule
\end{tabular}
\end{table}

\subsubsection{Baseline Model Ablation}
Although the performance of the baseline models is lower than that of NESA, models such as MLP still achieve reasonable performance. We present the ablated MLP model in Table \ref{table:baseline_ablation} and compare all its features to determine which feature contributes the most to the performance. We removed each feature one by one, and retrained the MLP model. We found that the MLP model, like NESA, largely depends on the context feature. It seems that MLP tends to choose empty slots based on the context features. 
% * <susanniekim@gmail.com> 2018-05-22T20:07:33.693Z:
% 
% > feature 
% depends on context features? (general plural)? 
% 
% ^ <dweller92@naver.com> 2018-05-23T06:42:27.915Z:
% 
% the feature.
%
% ^.
% * <susanniekim@gmail.com> 2018-05-21T19:38:35.725Z:
% 
% > feature
% no s or s?
% 
% ^ <krth32@gmail.com> 2018-05-22T12:05:45.795Z:
% 
% s. modified to "features"
%
% ^ <dweller92@naver.com> 2018-05-22T14:55:35.572Z.
% * <susanniekim@gmail.com> 2018-05-19T21:08:26.507Z:
% 
% > feature
% uc in MLP?
% 
% ^ <krth32@gmail.com> 2018-05-20T14:42:34.587Z:
% 
% added "in baseline models"
% 
% ^ <susanniekim@gmail.com> 2018-05-21T19:36:42.429Z:
% 
% what baseline models? sp
%
% ^ <krth32@gmail.com> 2018-05-22T12:09:31.240Z:
% 
% would "all the baseline models" be better?
%
% ^ <dweller92@naver.com> 2018-05-22T14:56:22.881Z.
% * <susanniekim@gmail.com> 2018-05-19T21:07:53.292Z:
% 
% > feature
% s no s? the no the?
% 
% ^ <krth32@gmail.com> 2018-05-22T12:11:56.966Z:
% 
% s, the. sorry i missed this
%
% ^ <dweller92@naver.com> 2018-05-22T14:56:25.985Z.
% * <susanniekim@gmail.com> 2018-05-19T21:06:23.060Z:
% 
% > MLP
% MLP or the MLP model?
% 
% ^ <krth32@gmail.com> 2018-05-20T14:45:13.364Z:
% 
% modified to "MLP model"
%
% ^ <susanniekim@gmail.com> 2018-05-21T19:37:02.824Z.

\begin{table*}
\small
\caption{Nearest Neighbors of Title/Intention Representations Given the Title \texttt{App project work} (duration 120 min.)}
\label{table:effect_intention}
\begin{tabular}{cccc}
\toprule
Title layer & \multicolumn{3}{c}{Intention layer} \\
\cmidrule{2-4}
& User A (Duration 120 min.) & \textbf{User B} (Duration 120 min.) & User A (\textbf{Duration 240 min.}) \\
\midrule
App project work (120) & Make V1 of \underline{app} (120) & Create paperwork for meetings (60) & Meet Databases Team (\underline{240}) \\
App work (540) & Do \underline{Databases} project (120) & \underline{Try} Fontana again (60) & App work (\underline{540}) \\
App Description to Richard (60) & \underline{Databases} (120) & \underline{Try} Peter @ UNK again (60) & Watch databases, do algorithmics (\underline{240}) \\
App w Goodman (60) & UNK and spot market (120) & \underline{Try} pepper Jaden Mark (60) & Databases Final Meeting (\underline{180}) \\
\bottomrule
\end{tabular}
\end{table*}

\begin{figure}[t]
\includegraphics[width=8.5cm, height=8.0cm]{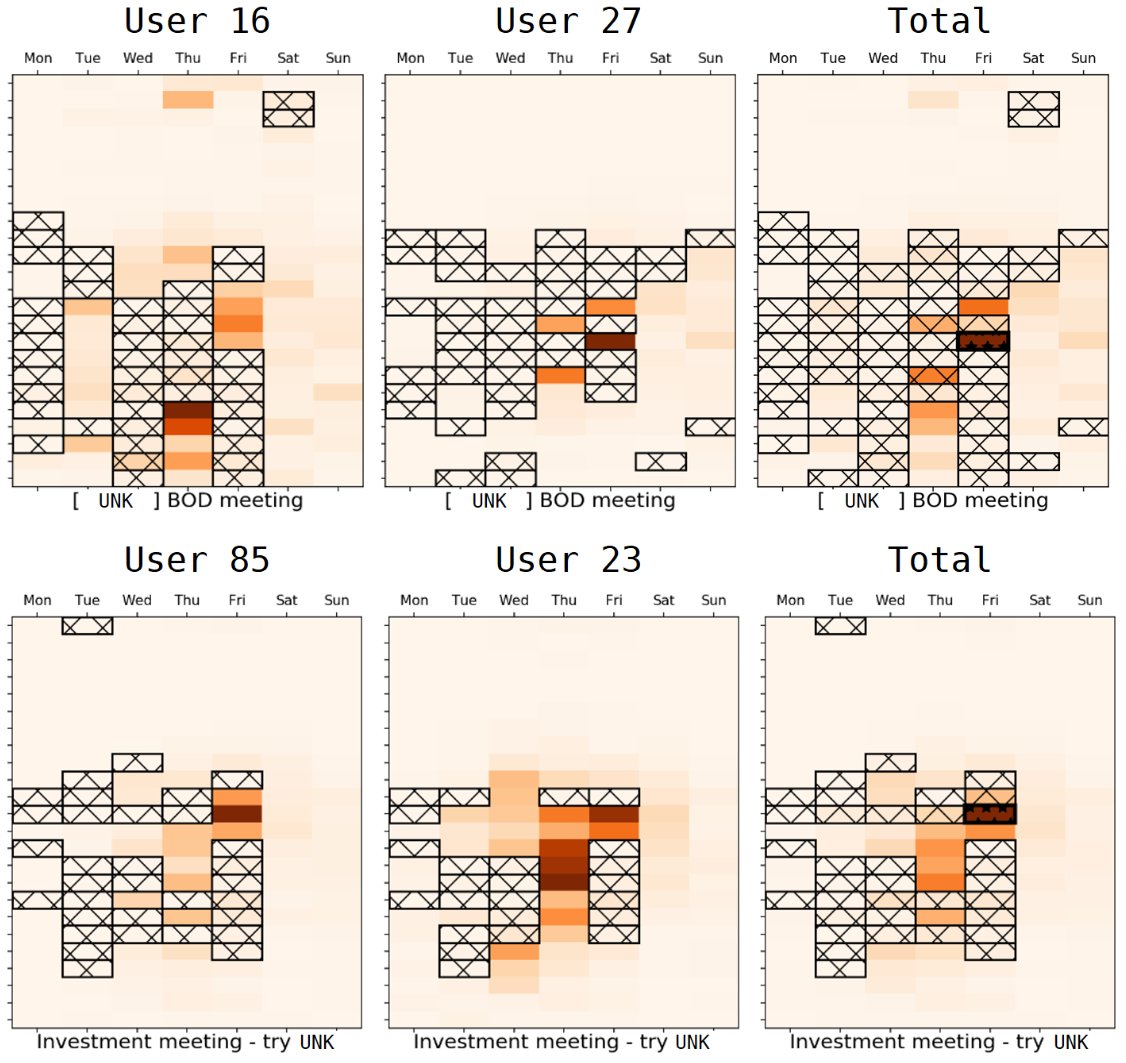}
\caption{Output probabilities of NESA in multi-attendee meeting scheduling.}
\label{fig:multi_meeting}
\end{figure}

\subsection{Qualitative Analysis}
\subsubsection{Effect of the Title Layer}
\indent Given different titles, NESA assigns different probabilities to each slot. In Figure \ref{fig:title_heatmap}, we visualized the output probabilities of NESA given four different input titles. The rows of each heatmap denote the hours in a day, and the columns denote the days in a week. The filled time slots are marked with lattices and the answers are marked with stars. For the title "Dinner with the people," NESA suggests mostly night times. Also, for the title "Late Lunch in Downtown," NESA not only suggests late lunch hours, but it also chooses days that the user may be in downtown. \textit{Workout} and \textit{Meeting} are more ambiguous keywords than \textit{Lunch} or \textit{Dinner}, but NESA suggests again suitable time slots based on each title. Figure \ref{fig:title_heatmap} shows \textit{workout} is done on weekends or at evening-time while \textit{Meetings} are held during office hours.\\
% * <susanniekim@gmail.com> 2018-05-19T21:20:41.487Z:
% 
% > Downtown
% why italics?
% 
% ^ <krth32@gmail.com> 2018-05-20T14:47:04.050Z:
% 
% in this paper, we italic event title words
%
% ^ <susanniekim@gmail.com> 2018-05-21T19:40:27.042Z:
% 
% but is Downtown an event? Downtown is more like a location, you write as a location too. 
%
% ^ <krth32@gmail.com> 2018-05-22T12:23:57.180Z:
% 
% it looks closer to a location. turned off italic
%
% ^ <dweller92@naver.com> 2018-05-22T14:56:42.545Z.
% * <susanniekim@gmail.com> 2018-05-19T21:17:56.956Z:
% 
% > for
% or to?
% 
% ^ <krth32@gmail.com> 2018-05-20T14:47:24.260Z:
% 
% modified to "to"
%
% ^ <susanniekim@gmail.com> 2018-05-21T19:40:45.313Z.
\indent In Table \ref{table:effect_title}, we show the 4 nearest neighbors of title representations of MLP and NESA. The distances between each representation were calculated using the cosine similarity. MLP's title representation is the element-wise average of word embeddings, and NESA uses the Title layer for title representations. With the title "Family lunch," we observe that MLP's title representations do not differentiate each keyword in event scheduling. Although the keyword \textit{lunch} should have more effect on event scheduling, most nearest neighbors of MLP's title representation are biased towards the keyword \textit{Family}, while nearest neighbors of NESA's title representation are mostly related to \textit{lunch}.

% \begin{figure}[t]
% \includegraphics[width=8.0cm]{figures/duration_emb_wide.png}
% \caption{Learned duration embeddings. Each point corresponds to minutes of duration.}
% \label{fig:dur_embedding}
% \end{figure}

\begin{figure}[t]
\includegraphics[width=8.5cm, height=8.0cm]{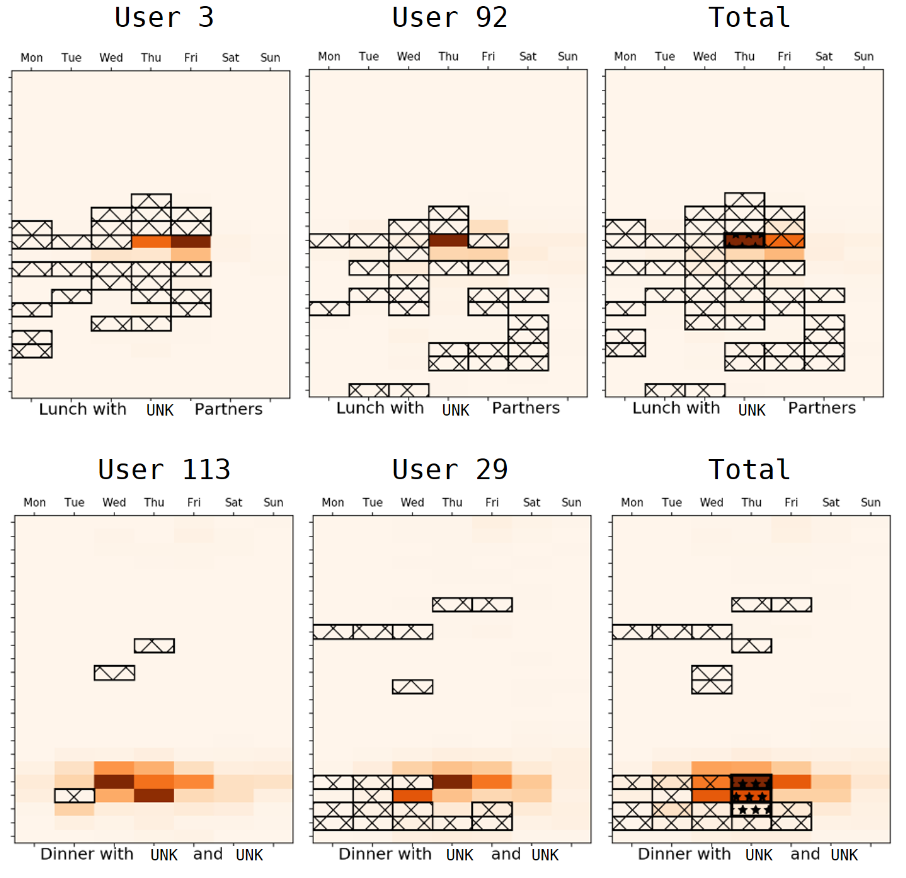}
\caption{Output probabilities of NESA in multi-attendee event scheduling given lunch and dinner events.}
\label{fig:multi_life}
\end{figure}

\begin{figure}[t]
\includegraphics[width=8.5cm, height=8.0cm]{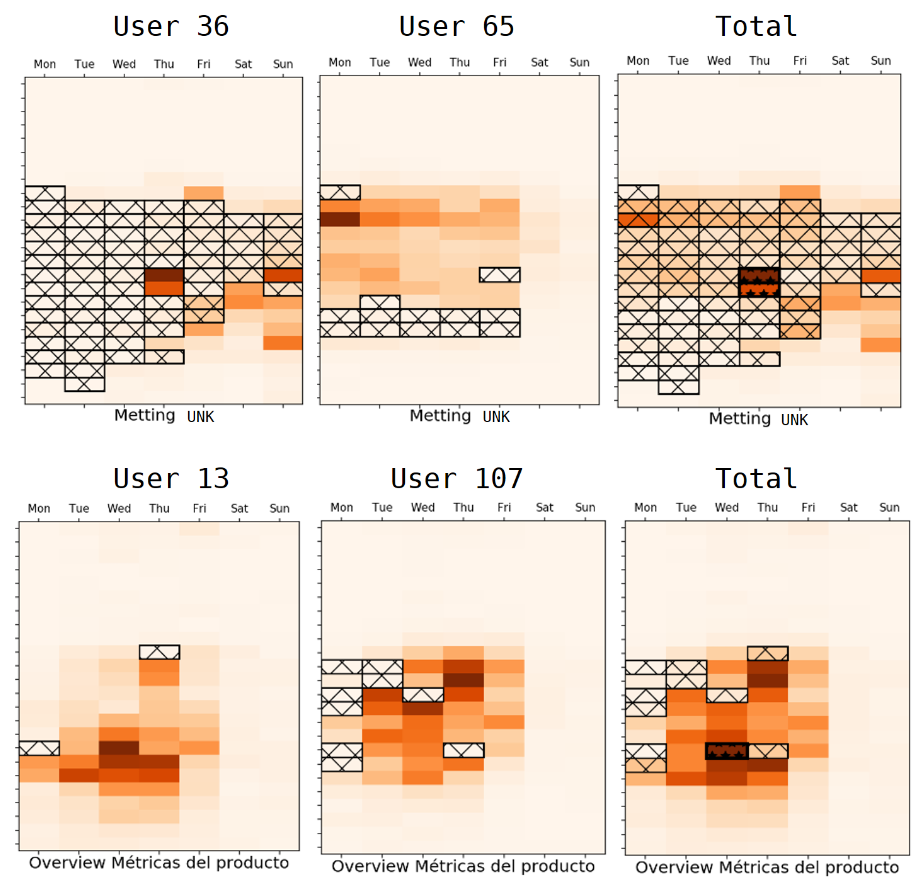}
\caption{Output probabilities of NESA in multi-attendee event scheduling given misspelled and non-English events.}
% * <susanniekim@gmail.com> 2018-05-22T20:15:47.604Z:
% 
% > misspelled
% misspelled events? or misspellings?  
% 
% ^.
\label{fig:multi_char}
\end{figure}

\subsubsection{Effect of the Intention Layer}
The Intention layer in NESA combines different types of attributes from calendar data. In Table \ref{table:effect_intention}, we present the 4 nearest neighbors of the title and intention representations based on the cosine similarities. Given the title "App project work," the Title layer simply captures semantic representations of the title. Titles with similar meanings such as "App work" are its nearest neighbors (1st column). On the other hand, the nearest neighbors of the intention representation are related to not only the keyword \textit{app} but also the keyword \textit{database}, which is one of user A's frequent terms (2nd column). We observe that the intention representation changes by replacing user A with user B who frequently uses the term \textit{Try} (3rd column). The duration attribute is also well represented as events with longer durations are closer to user A's 240 minute long event (4th column).

\subsubsection{Multi-Attendee Event Scheduling Analysis}
In Figures \ref{fig:multi_meeting}--\ref{fig:multi_char}, we present examples of multi-attendee event scheduling. Using NESA, we obtain each user's preferred time slots, and the suggested time slots for multi-attendee events are calculated by Equation \ref{eq:multi}. Again, the filled time slots are marked with lattices and the answers are marked with stars. We show a multi-attendee event in each row, and each row contains the preferences of two different users and their summed preference (total). We anonymized any pronouns as \textit{UNK} tokens for privacy issues. \\
\indent Figure \ref{fig:multi_meeting} shows examples of event scheduling for meetings. The two examples clearly show that NESA understands each user's calendar context, and suggests time intervals mostly during office hours. Figure \ref{fig:multi_life} shows appointments such as lunch and dinner rather than meetings. While each example accurately represents the purpose of each event, note that NESA does not suggest weekends for "Lunch with UNK Partners." We think that NESA understands the keyword \textit{Partner}, which is frequently related to formal meetings. In Figure \ref{fig:multi_char}, we show how misspellings (\textit{Metting} for \textit{meeting}) and non-English ("M\'{e}tricas del producto" means "product's metric" in Spanish) are understood by NESA. As NESA has the Title layer that leverages the characters of infrequent words, NESA successfully suggests suitable office hours for each event.

\section{Conclusions and Future Work}
In this paper, we proposed a novel way to fully make use of raw online calendar data for event scheduling. Our proposed model NESA learns how to perform event scheduling directly from raw calendar data, and to consider user preferences and calendar contexts. We also showed that deep neural networks are highly effective in scheduling events. Unlike previous works, we leveraged a large-scale online calendar dataset in the Internet standard format, which makes our approach more applicable to other systems. NESA achieves the best performance among the existing baseline models on both personal and multi-attendee event scheduling tasks. \\
% * <susanniekim@gmail.com> 2018-05-19T22:19:33.554Z:
% 
% > based on 
% which uses the format?  in the?
% 
% ^ <krth32@gmail.com> 2018-05-20T15:06:45.652Z:
% 
% modified to "in"
%
% ^ <susanniekim@gmail.com> 2018-05-21T20:07:20.307Z.
% * <susanniekim@gmail.com> 2018-05-19T22:14:45.530Z:
% 
% > learns event scheduling directly from raw user calendars
% learns how to/ performs event sched ...using cal data?  
% 
% ^ <krth32@gmail.com> 2018-05-20T15:07:34.310Z:
% 
% ***modified to "learns how to perform event scheduling directly using raw calendar data"
% 
% ^ <susanniekim@gmail.com> 2018-05-21T20:08:04.683Z.
\indent For future work, we plan to study the relationships between users for multi-attendee event scheduling. Unfortunately, such relationship information is not provided in the standard calendar format, and should be inferred from multi-attendee event scheduling examples. Once we obtain more multi-attendee calendar events, such an approach would produce more sophisticated multi-attendee scheduling systems.
% * <susanniekim@gmail.com> 2018-05-19T22:22:52.809Z:
% 
% > formats
% s no s? 
% 
% ^ <krth32@gmail.com> 2018-05-20T15:06:20.151Z:
% 
% no s
%
% ^ <susanniekim@gmail.com> 2018-05-21T19:23:53.093Z.
% * <susanniekim@gmail.com> 2018-05-19T22:22:21.908Z:
% 
% > consider
% or study? or use? 
% 
% ^ <krth32@gmail.com> 2018-05-20T15:06:12.573Z:
% 
% modified to "study"
%
% ^ <susanniekim@gmail.com> 2018-05-21T19:24:06.502Z.

\begin{acks}
This research was supported by National Research Foundation of Korea (NRF-2017R1A2A1A17069645, NRF-2017M3C4A7065887).
\end{acks}

\bibliographystyle{sample-sigconf}
\bibliography{sample-sigconf}

\end{document}